\documentclass[11pt,a4paper]{article}
\usepackage[hyperref]{acl2019}
\usepackage{times}
\usepackage{latexsym}

\usepackage{url}

\aclfinalcopy 
\def\aclpaperid{1455} 

\usepackage{microtype}

\usepackage{qtree}
\usepackage{forest}

\usepackage{float}
\usepackage{graphicx} 
\usepackage{amsthm, amsmath}
\usepackage{amssymb}
\usepackage{latexsym}

\usepackage{algpseudocode}
\usepackage{algorithm}

\usepackage{booktabs}

\usepackage{caption}
\usepackage{subcaption}

\usepackage{bussproofs} 
\usepackage{stackengine}

\newcommand{\sortof}[1]{`#1'}
\newcommand{\of}[1]{\left(#1\right)}

\newlength\mylen

\makeatletter
\def\th@definition{%
  \thm@notefont{}
  \normalfont 
}
\makeatother

\theoremstyle{definition} 

\theoremstyle{definition}

\usepackage{amssymb}

\usepackage{pifont}

\usepackage[pdf,singlefile]{graphviz}

\newcommand{\depscl}{0.35}

\usepackage{setspace}

\usepackage{bm}
\usepackage{xspace}
\usepackage[utf8]{inputenc}

\newcommand{\app}[1]{\text{\textsc{App}}\ensuremath{_{#1}}}  
\newcommand{\modify}[1]{\text{\textsc{mod}}\ensuremath{_{#1}}\xspace}

\newcommand{\src}[1]{\text{\textsc{#1}}\xspace}                                  
\newcommand{\obj}[1][]{\text{\src{o}}\ensuremath{_{#1}}}               

\newcommand{\G}[1]{\ensuremath{G_{\text{#1}}}} 

\newcommand{\el}[1]{\text{`#1'}}
\newcommand{\nl}[1]{\nodelabel{#1}}
 \newcommand{\word}[1]{``#1''}
 \newcommand{\nodelabel}[1]{``#1''}
 
\usepackage{tikz}
\usepackage{tikz-qtree}

\usetikzlibrary{graphs}
\usetikzlibrary{shapes,arrows}
\usetikzlibrary{positioning}
\usetikzlibrary{quotes}

\tikzset{snode/.style={
    ellipse,
    minimum size=6mm,
    very thick,
    draw=black,
    font=\rmfamily}}

\tikzset{ssrc/.style={
    ellipse,
    minimum size=6mm,
    very thick,
    fill=black!10,
    draw=black,
    text=black,
    font=\rmfamily}}

\tikzset{sanno/.style={node distance=-1mm,font=\rmfamily\small}}

\tikzset{sedgel/.style={
    color=black,
    sloped, above,
    font=\rmfamily\small}}

\tikzset{sedge/.style={
    thick,
    >=stealth'
}}

\usepackage{array}

\newcommand{\xmark}{\ding{55} }
\newcommand{\cin}[1]{\scalebox{0.8}{\ensuremath{\pm #1}}}

\newcommand{\wirACL}{groschwitz18:_amr_depen_parsin_typed_seman_algeb}

\newcommand\blfootnote[1]{%
  \begingroup
  \renewcommand\thefootnote{}\footnote{#1}%
  \addtocounter{footnote}{-1}%
  \endgroup
}

\title{Compositional Semantic Parsing Across Graphbanks} 

\author{Matthias Lindemann$^{*}$ \and Jonas Groschwitz$^{*}$  \and Alexander Koller\\
  Department of Language Science and Technology\\
  Saarland University\\
  \url{{mlinde|jonasg|koller}@coli.uni-saarland.de}}

\date{}

\begin{document}
\maketitle
\begin{abstract}
  
Most semantic parsers that map sentences to graph-based meaning
representations are hand-designed for specific graphbanks. We present
a compositional neural semantic parser which achieves, for the first
time, competitive accuracies across a diverse range of
graphbanks. Incorporating BERT embeddings and multi-task learning
improves the accuracy further, setting new states of the art on DM,
PAS, PSD, AMR 2015 and EDS.


  \blfootnote{\hspace{-1.5mm}$^{*}$Equal contribution}%
\end{abstract}

\section{Introduction} \label{sec:introduction}

Over the past few years, a wide variety of semantic \emph{graphbanks}
have become available. Although these corpora all pair
natural-language sentences with graph-based semantic representations,
they differ greatly in the design of these graphs
\cite{kuhlmann16:_towar_catal_linguis_graph_banks}. Some, in
particular the DM, PAS, and PSD corpora of the SemEval shared task on
Semantic Dependency Parsing \cite{OepenKMZCFHU15}, use the tokens of
the sentence as nodes and connect them with semantic relations.  By
contrast, the AMRBank \cite{amBanarescuBCGGHKKPS13} represents the
meaning of each word using a nontrivial concept graph; the EDS
graphbank \cite{OpenSDP}
encodes
MRS representations \citep{copestake2005mrs} as graphs with a many-to-many relation
between tokens and nodes. In EDS, graph nodes are explicitly aligned
with the tokens; in AMR, the alignments are implicit. The graphbanks
also exhibit structural differences in their modeling of
e.g.~coordination or copula.

Because of these differences in annotation schemes, the best
performing semantic parsers are typically designed for one or very few
specific graphbanks. For instance, the currently best system for DM,
PAS, and PSD \cite{Dozat18SemanticDependency} assumes dependency
graphs and cannot be directly applied to EDS or AMR. Conversely, top
AMR parsers \cite{LyuTitov18AMR} invest heavily into identifying
AMR-specific alignments and concepts, which may not be useful in other
graphbanks.  \citet{Hershcovich18} parse across different
semantic graphbanks (UCCA, DM, AMR), but focus on UCCA and do poorly on DM. The
system of \newcite{BuysBlunsom17} set a state of the art on EDS at the
time, but does poorly on AMR.

In this paper, we present a single semantic parser that does very well
across all of DM, PAS, PSD, EDS and AMR (2015 and 2017). Our system is
based on the compositional neural AMR parser of
\newcite{groschwitz18:_amr_depen_parsin_typed_seman_algeb}, which
represents each graph with its compositional tree structure and learns
to predict it through neural dependency parsing and supertagging. We
show how to heuristically compute the latent compositional structures
of the graphs of DM, PAS, PSD, and EDS. This base parser already
performs near the state of the art across all six graphbanks. We
improve it further by using pretrained BERT embeddings
\cite{devlin2018bert} and multi-task learning. With this, we set new
states of the art on DM, PAS, PSD, AMR 2015, as well as (among systems
that do not use specialized knowledge about the corpus) on EDS.


\section{Semantic parsing with the AM algebra} \label{sec:amalgebra}

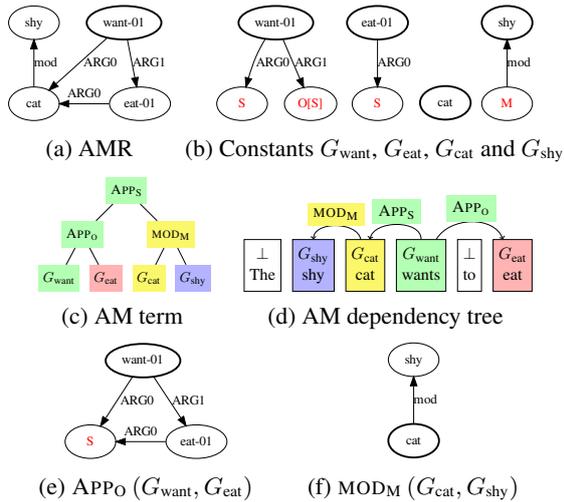
\begin{figure}
    \centering
    \begin{subfigure}[b]{0.3\linewidth}
        \centering
	    \includegraphics[scale=0.35]{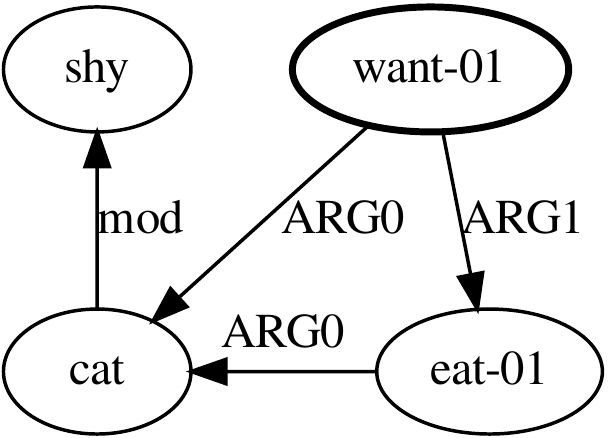}
		\caption{AMR}\label{fig:amr-am:amr}
	\end{subfigure}
	\begin{subfigure}[b]{0.65\linewidth}
        \centering
	    \includegraphics[scale=0.35]{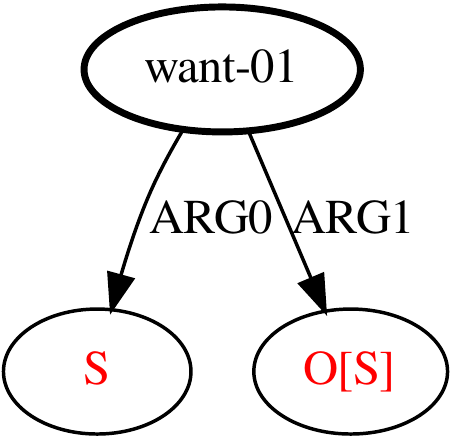}
	    \includegraphics[scale=0.35]{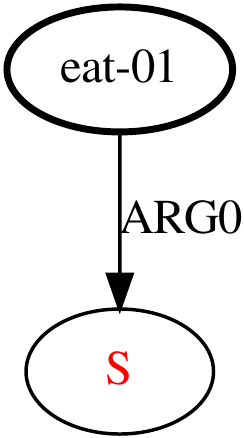}
	    \includegraphics[scale=0.35]{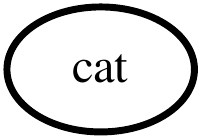}
	    \includegraphics[scale=0.35]{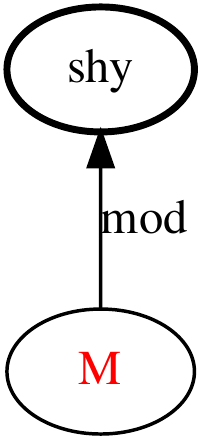}
		\caption{Constants \G{want}, \G{eat}, \G{cat} and \G{shy}}\label{fig:amr-am:consts}
	\end{subfigure}\\[3pt]

	\begin{subfigure}[b]{0.3\linewidth}
		\begin{center}
			\resizebox{\linewidth}{!}{
			\begin{forest}
				for tree={align=center}
				[{\app{\src{S}}}, for tree={fill=green!30}
					[{\app{\obj}}
						[{\G{want}}]
						[{\G{eat}}, for tree={fill=red!30}]
					]
					[{\modify{\src{M}}}, for tree={fill=brown!10!yellow!60}
						[{\G{cat}}]
						[{\G{shy}}, for tree={fill=blue!30}]
					]
				]
			\end{forest}}
		\end{center}
		\vspace{-6pt}
		\caption{AM term}\label{fig:amr-am:term}
	\end{subfigure}\quad
		\begin{subfigure}[b]{0.5\linewidth}
		\begin{center}
			\resizebox{\linewidth}{!}{
			\begin{forest}
				for tree={draw,align=center, no edge,l sep =0pt, l=\baselineskip}
				[parent, phantom
				[{$\bot$\\The}, name=the]
				[{\G{shy}\\shy}, name=shy, fill=blue!30]
				[{\G{cat}\\cat}, name=cat, fill=brown!10!yellow!60]
				[{\G{want}\\wants}, name=want, fill=green!30]
				[{$\bot$\\to}, name=to]
				[{\G{eat}\\eat}, name=eat, fill=red!30]
				]
				\draw[->] (want) to[out=north, in=80, edge node={node[above, fill=green!30] {\app{\src{S}}}}] (cat);
				\draw[->] (want) to[out=60, in=110, edge node={node[above, fill=green!30] {\app{\obj}}}] (eat);
				\draw[->] (cat) to[ out=100, in=north, edge node={node[above, fill=brown!10!yellow!60] {\modify{\src{M}}}}] (shy);
			\end{forest}}
		\end{center}
		\vspace{-6pt}
		\caption{AM dependency tree}\label{fig:amr-am:deptree}
	\end{subfigure}\\[3pt]
	
	\begin{subfigure}[b]{0.4\linewidth}
		\centering
		\includegraphics[scale=0.35]{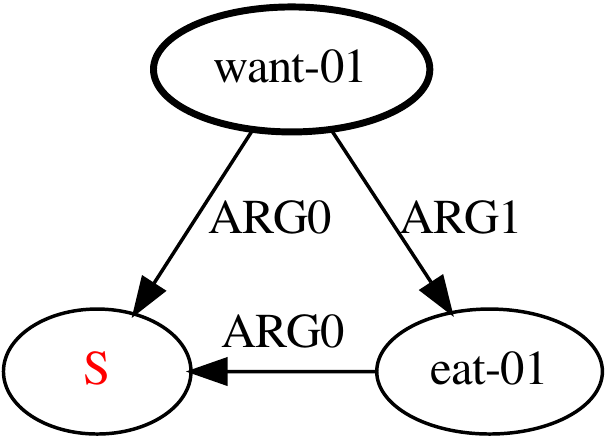}
		\caption{$\app{\src{O}}\of{\G{want},\G{eat}}$}\label{fig:amr-am:wantEat}
	\end{subfigure}$\:\:\:\:$
	\begin{subfigure}[b]{0.4\linewidth}
		\centering
		\includegraphics[scale=0.35]{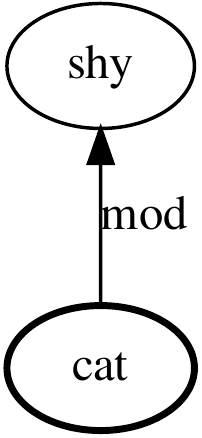}
		\caption{$\modify{\src{M}}\of{\G{cat},\G{shy}}$}\label{fig:amr-am:shyCat}
	\end{subfigure}
\vspace{-3pt}
    \caption{AMR for \textit{The shy cat wants to eat} with its AM
      analysis.}
    \label{fig:amr-am}
\end{figure}

The \emph{Apply-Modify (AM) Algebra} \cite{graph-algebra-17, GroschwitzDiss} builds
graphs from smaller graph fragments called
\emph{as-graphs}. Fig.~\ref{fig:amr-am:consts} shows some as-graphs
from which the AMR in Fig.~\ref{fig:amr-am:amr} can be
constructed. Take for example the graph $\G{want}$. Some of its nodes
are marked with red \emph{sources}, here \src{S} and \src{O}. These
represent \sortof{argument slots} to be filled. The \src{O}-source in
$\G{want}$ is \emph{annotated} with \emph{type} $[\src{S}]$, which
will be explained below. Further, in each as-graph, one node is marked
as a special \emph{root source}, drawn here with a bold outline.

There are two operations in the AM Algebra that combine
as-graphs. First, the \emph{apply} operation \app{\src{X}} for a
source \src{X}, as in $\app{\src{O}}\of{\G{want},\G{eat}}$ with the
result shown in Fig.~\ref{fig:amr-am:wantEat}. The operation combines two
as-graphs, a head and an argument, by filling the head's
\src{X}-source with the root of the argument.
Nodes in both graphs with the same source are unified,
i.e.~here the two nodes marked with an \src{S}-source become one node. The type annotation $[\src{S}]$ at the $\src{O}$-source
of \G{want} requests the argument to have an \src{S}-source (which
\G{eat} has). If the argument does not fulfill the request at a source's annotation, the operation is not
\emph{well-typed} and thus not allowed. 

The second operation is \emph{modify}, as in
$\modify{\src{M}}\of{\G{cat},\G{shy}}$ with the result shown in
Fig.~\ref{fig:amr-am:shyCat}. Here, \G{cat} is the head and \G{shy}
the modifier, and in the operation, \G{shy} attaches with its \src{M}
source at \G{cat} and loses its own root. We obtain the final graph
with the \app{\src{S}} operation at the top of the term in
Fig.~\ref{fig:amr-am:term}, combining the two partial results we have built
so far.

\captionsetup[sub]{font=small}

\textbf{AM dependency parsing.} By tracking the ``semantic heads'' of
each subtree of an AM term as in Fig.~\ref{fig:amr-am}c, we can encode
AM terms as \emph{AM dependency trees} (Fig.~\ref{fig:amr-am}d):
whenever the AM term combines two graphs with some operation, we add
a dependency edge from one semantic head to the other
\cite{groschwitz18:_amr_depen_parsin_typed_seman_algeb}.

We can then parse a sentence into a graph by predicting an as-graph
(or the absence of one, written `$\bot$') for each token in the sentence, along with a
well-typed AM dependency tree that connects them. This AM dependency
tree evaluates deterministically to a
graph. \newcite{groschwitz18:_amr_depen_parsin_typed_seman_algeb} show
how to perform accurate AMR parsing by training a neural supertagger
to predict as-graphs for the words and a neural dependency (tree)
parser to predict the AM dependency trees. Here we use their basic
models for predicting edge and supertag scores. Computing the
highest-scoring well-typed AM dependency is NP-complete; we use
Groschwitz et al.'s fixed-tree parser to compute it approximatively.


\section{Decomposing the graphbanks}

A central challenge with AM dependency parsing is that the AM
dependency trees in the training corpus are latent: Strings are
annotated with graphs (Fig.~\ref{fig:amr-am:amr}), but we need the
supertags and AM dependency trees (Fig.~\ref{fig:amr-am:deptree}).

\begin{figure}[ht] 
    
    \rotatebox{90}{
        \begin{minipage}[c][][c]{40pt}
        \subcaption{DM}\label{fig:dm}
        \end{minipage}
	}
    \begin{minipage}[c][][c]{0.45\textwidth}
        \begin{center}
        \scalebox{0.87}{
		    \small
			\begin{forest}
				for tree={align=center, no edge,l sep =0pt, l=\baselineskip}
				[parent, phantom
					[{The}, name=the]
					[{shy}, name=shy]
					[{cat}, name=cat]
					[{wants}, name=want]
					[{to}, name=to]
					[{eat}, name=eat]
				]
				\draw[->] (the) to[pos=0.35, out=90, in=100, edge node={node[above] {\tiny BV}}] (cat);
				\draw[->] (shy) to[pos=0.2,out=90, in=110, edge node={node[above] {\tiny ARG1}}] (cat);
				\draw[->] (want) to[pos=0.35,out=90, in=80, edge node={node[above] {\tiny ARG1}}] (cat);
				\draw[->] (want) to[out=90, in=100, edge node={node[below] {\tiny ARG2}}] (eat);
				\draw[->] (eat) to[out=90, in=90, edge node={node[above] {\tiny ARG1}}] (cat);
			\end{forest}}
		    \hrule
		    \scalebox{0.87}{
		    \input{texedpics/dmdep.tex}}
		\end{center}
	\end{minipage}
	\\[3pt]\noindent\rule{0.49\textwidth}{0.4mm}

    \rotatebox{90}{
        \begin{minipage}[c][][c]{40pt}
        \subcaption{PAS}\label{fig:pas}
        \end{minipage}
	}
    \begin{minipage}[c][][c]{0.45\textwidth}
        \begin{center}
        \scalebox{0.87}{
		    \small
			\begin{forest}
				for tree={align=center, no edge,l sep =0pt, l=\baselineskip}
				[parent, phantom
					[{The}, name=the]
					[{shy}, name=shy]
					[{cat}, name=cat]
					[{wants}, name=want]
					[{to}, name=to]
					[{eat}, name=eat]
				]
				\draw[->] (the) to[pos=0.5, out=90, in=100, edge node={node[above] {\tiny det\_ARG1}}] (cat);
				\draw[->] (shy) to[pos=0.0,out=90, in=110, edge node={node[above=3pt] {\tiny adj\_ARG1}}] (cat);
				\draw[->] (want) to[pos=0.35,out=90, in=80, edge node={node[above] {\tiny verb\_ARG1}}] (cat);
				\draw[->] (want) to[out=90, in=100, edge node={node[above=-1pt] {\tiny verb\_ARG2}}] (eat);
				\draw[->] (to) to[pos=0.1, out=60, in=120, edge node={node[above] {\tiny comp\_ARG1}}] (eat);
				\draw[->] (eat) to[out=70, in=90, edge node={node[above] {\tiny verb\_ARG1}}] (cat);
			\end{forest}}
		    \hrule
		    \hspace{-5pt}
		    \scalebox{0.87}{
		    \input{texedpics/pasdep.tex}}
		\end{center}
	\end{minipage}
	\\[3pt]\noindent\rule{0.49\textwidth}{0.4mm}

    \rotatebox{90}{
        \begin{minipage}[c][][c]{40pt}
        \subcaption{PSD}\label{fig:psd}
        \end{minipage}
	}
    \begin{minipage}[c][][c]{0.45\textwidth}
        \begin{center}
        \scalebox{0.87}{
		    \small
			\begin{forest}
				for tree={align=center, no edge,l sep =0pt, l=\baselineskip}
				[parent, phantom
					[{The}, name=the]
					[{shy}, name=shy]
					[{cat}, name=cat]
					[{wants}, name=want]
					[{to}, name=to]
					[{eat}, name=eat]
				]
				\draw[->] (cat) to[pos=0.8, out=110, in=90, edge node={node[above] {\tiny RSTR}}] (shy);
				\draw[->] (want) to[pos=0.35,out=90, in=80, edge node={node[above] {\tiny ACT-arg}}] (cat);
				\draw[->] (want) to[out=90, in=100, edge node={node[below] {\tiny PAT-arg}}] (eat);
				\draw[->] (eat) to[out=90, in=90, edge node={node[above] {\tiny ACT-arg}}] (cat);
			\end{forest}}
		    \hrule
		    \scalebox{0.87}{
		    \input{texedpics/psddep.tex}}
		\end{center}
	\end{minipage}
	\\[3pt]\noindent\rule{0.49\textwidth}{0.4mm}

    \rotatebox{90}{
        \begin{minipage}[c][][c]{40pt}
        \subcaption{EDS}\label{fig:eds}
        \end{minipage}
	}
    \begin{minipage}[c][][c]{0.45\textwidth}
        \begin{center}
		    \includegraphics[scale=0.38]{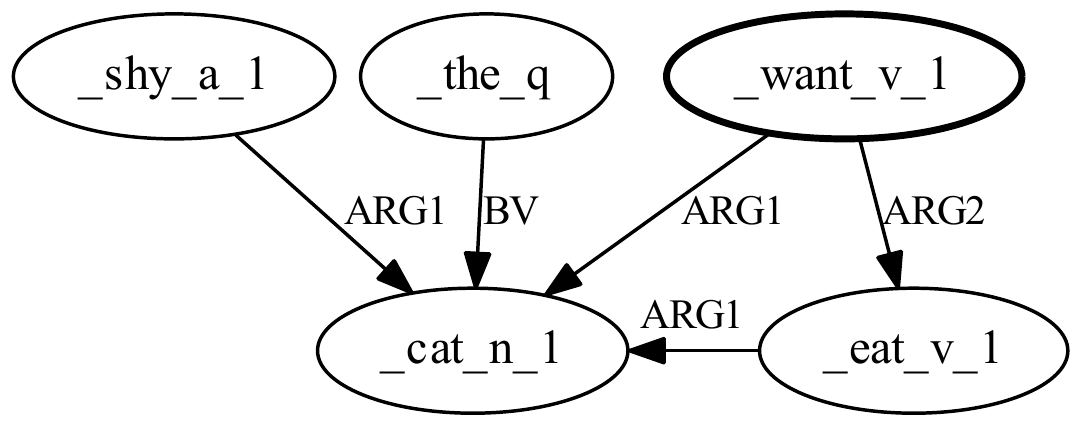}
		    \hrule
		    \hspace{-10pt}
		    \scalebox{0.87}{
		    \input{texedpics/edsdep.tex}}
		\end{center}
	\end{minipage}
	\vspace{-6pt}
    \caption{Semantic representations for \textit{The shy cat wants to eat}, each with an AM dependency tree below.}
    \label{fig:many}
\end{figure}

\citet{groschwitz18:_amr_depen_parsin_typed_seman_algeb} describe a
heuristic algorithm to obtain AM dependency trees for AMRs (\emph{decomposition}). They first
\textbf{align} each node in the graph with a word token; then
\textbf{group} the edges together with either their source or target
nodes, depending on the edge label; choose a \textbf{source name} for
the open slot at the other end of each attached edge; and match
reentrancy patterns to determine \textbf{annotations} for each source. The dependency edges follow from these decisions.

Groschwitz et al.\ worked these steps out only for AMR. Here we extend
their work to DM, PAS, PSD, and EDS (see Figure~\ref{fig:many}); this
is the central technical contribution of this paper.

\subsection{The graphbanks}
Before we discuss the decomposition process, let us examine the key
similarities and differences of AMR, DM, PAS, PSD and EDS. Most
obvious is that DM, PAS and PSD are dependency graphs
(Figure~\ref{fig:many}a-c) where the nodes of the graphs are the words
of the sentences, while EDS (Figure~\ref{fig:many}d) and AMR use nodes
related to, but separate from the words. Node-to-word alignments are
given in EDS, but not in AMR, where predicting them is hard
\cite{LyuTitov18AMR}.

\begin{figure}
    \centering
    \begin{subfigure}[b]{0.4\linewidth}
        \centering
        \begin{tikzpicture}
        \node[](amr) at (0,0){\includegraphics[scale=0.4]{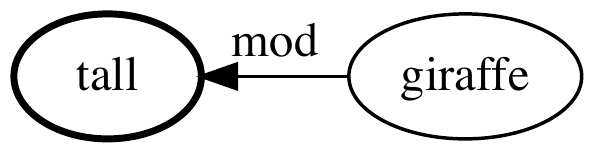}};
        \node[](a) at (-35pt,20pt){\small (a)};
        \end{tikzpicture}
	\end{subfigure}
	\begin{subfigure}[b]{0.5\linewidth}
        \centering
        \small
			\begin{forest}
				for tree={align=center, no edge,l sep =0pt, l=\baselineskip}
				[parent, phantom
					[{Giraffes}, name=giraffe]
					[{are}, name=are]
					[{tall}, name=tall]
				]
				\draw[->] (tall) to[pos=0.5, out=70, in=120, edge node={node[above=-1pt] {\tiny adj\_ARG1}}] (giraffe);
				\draw[->] (are) to[pos=0.4,out=110, in=70, edge node={node[above=0.5pt] {\tiny verb\_ARG1}}] (giraffe);
				\draw[->] (are) to[pos=0.3,out=70, in=110, edge node={node[above=-1pt] {\tiny verb\_ARG2}}] (tall);
				\node[](b) at (-55pt,10pt){(b)};
			\end{forest}
	\end{subfigure}
	
    \caption{AMR (a) and PAS (b) for \textit{Giraffes are tall}.}
    \label{fig:copula}
\end{figure}
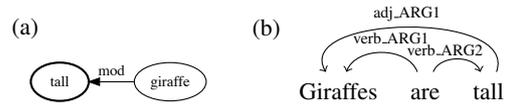

In all graphbanks we consider here, the edges express semantic
relations between the nodes. Several similarities exist: in our
example, all graphbanks have edges from \word{wants} and \word{eat} to \word{cat} that indicate that the cat is both the
wanter and the eater. These are for example the \el{ARG0} edges in AMR
and the \el{ACT-arg} edges in PSD. In fact, all five graphs show a
triangle structure between \word{want}, \word{eat} and \word{cat} that
is characteristic of control verbs. Similarly, all graphs have an edge
indicating that \word{shy} modifies \word{cat}, although edge label
and edge direction vary. However, the graphbanks
differ not only in edge directions and labels, but also structurally.
For example, DM, PAS and EDS annotate determiners while AMR and PSD do not.
Figure~\ref{fig:copula} shows a reentrancy structure for a copular \word{are} in PAS that
is not present in AMR.

\captionsetup[sub]{font=small}

\subsection{Our decomposition method} \label{sec:our-decomp-meth}
We adapt the decomposition procedure of Groschwitz et al.
in the following ways. We sketch the most interesting
points here; full details are in the supplementary materials.

\textbf{Alignments} are given in EDS and not necessary in DM, PAS, and PSD.

\textbf{Grouping.} We follow two principles in grouping edges with
nodes: Edges between heads and arguments always belong with the head, and
edges between heads and modifiers with the modifier (regardless
of the direction into which the edge points). This yields
supertags that generalize well, e.g.~a noun has the same supertag no
matter whether it has a determiner, whether it is modified by
adjectives, whether is agent, and so on.

We find that for all graphbanks,
just knowing the edge label is enough to group an edge properly.
Thus, we manually decide for each
of the 216 edge labels of all graphbanks whether the edges with this
label are to be grouped with their target or source node. 
For instance, \el{ACT-arg} edges in PSD and
\el{verb\_ARG1} edges in PAS are argument-type edges grouped with
their source node (they point from a verb to its agent). \el{RSTR} edges in PSD and \el{adj\_ARG1} edges in PAS
are modifier-type and grouped with the adjective; the former is grouped with its target node and the
latter with its source.
In DM, \el{ARG1} edges can
be both modifier- or argument-type (they are used for both adjectives and 
verbs); grouping them with their source
node is the correct choice in both cases.

\textbf{Source names.} We largely reuse Groschwitz et al.'s source
names, which are loosely inspired by (deep) syntactic relations, and
map the edge labels of each graphbank to preferred source names.  For
example, in PSD we associate \el{ACT-arg} edges with \src{S} sources
(for ``subject''). Some source names are new, such as \src{D} for
determiners in DM, PAS and EDS (AMRs do not represent determiners).

\textbf{Annotations.}
Groschwitz et al.'s algorithm for assigning annotations to sources
carries over to the other graphbanks.
For patterns that are the same across all graphbanks,
such as the \sortof{triangle} created by the control verb \word{want} in
Figures~\ref{fig:amr-am} and~\ref{fig:many}, we can re-use the same
pattern as for AMR. Thus, control verbs are identified automatically, and their sources
are assigned annotations which enforce the appropriate argument sharing.

Interestingly, the original patterns are
useful beyond their initial design. We found that for phenomena that cause reentrancies in the new graphbanks, but not in AMR -- such as copula in PAS, c.f.~Figure~\ref{fig:copula} --
there was typically a suitable pattern designed for a different phenomenon in AMR. E.g.~for
copula in PAS, the control pattern works.

We thus only update patterns that depend on edge labels; for
instance, coordinations in PAS are characterized through their
\el{coord\_ARGx} edges.

\begin{figure}
\centering
		\begin{center}
		\small
			\begin{forest}
				for tree={align=center, no edge,l sep =0pt, s sep=0pt, l=\baselineskip, inner sep=0pt, outer sep = 0.5pt}
				[parent, phantom
					[{John}, baseline, name=john]
					[{and}, name=and]
					[{Mary}, name=mary]
					[{sing}, name=sing]
				]
				\draw[->] (john) to[pos=0.75, out=60, in=110, edge node={node[above] {\tiny and\_c}}] (mary);
				\draw[->] (sing) to[out=110, in=80, edge node={node[above] {\tiny ARG1}}] (john);
				\node[](a) at (-28pt,17pt){(a)};
			\end{forest}
			\hspace{-10pt}
			\begin{forest}
				for tree={align=center, no edge,l sep =0pt, s sep=0pt, l=\baselineskip, inner sep=0pt, outer sep = 0.5pt}
				[parent, phantom
					[{John}, baseline, name=john]
					[{and}, name=and]
					[{Mary}, name=mary]
					[{sing}, name=sing]
				]
				\draw[->] (and) to[pos=0.75, out=270, in=270, edge node={node[below=5pt] {\tiny CONJ.member}}] (mary);
				\draw[->] (and) to[pos=0.75, out=270, in=270, edge node={node[below] {\tiny CONJ.member}}] (john);
				\draw[->] (sing) to[out=90, in=90, edge node={node[above] {\tiny ACT-arg}}] (john);
				\draw[->] (sing) to[pos=1.0, out=90, in=90, edge node={node[above=3.5pt] {\tiny ACT-arg}}] (mary);
				\node[](b) at (-28pt,17pt){(b)};
			\end{forest}
			\hspace{-10pt}
			\begin{forest}
				for tree={align=center, no edge,l sep =0pt, s sep=0pt, l=\baselineskip, inner sep=0pt, outer sep = 0.5pt}
				[parent, phantom
					[{John}, baseline, name=john]
					[{and}, name=and]
					[{Mary}, name=mary]
					[{sing}, name=sing]
				]
				\draw[->] (and) to[pos=0.75, out=270, in=270, edge node={node[below=5pt] {\tiny CONJ.member}}] (mary);
				\draw[->] (and) to[pos=0.75, out=270, in=270, edge node={node[below] {\tiny CONJ.member}}] (john);
				\draw[->] (sing) to[out=90, in=90, edge node={node[above] {\tiny ACT-arg}}] (and);
				\node[](c) at (-28pt,17pt){(c)};
			\end{forest}
		\end{center}\vspace{-5pt}
	\caption{Coordination in (a) DM and (b, c) PSD.}\label{fig:coord}
\end{figure}
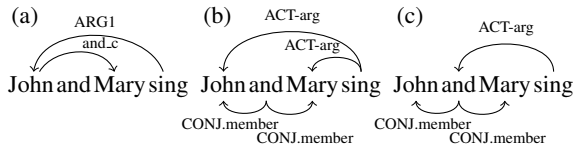

\textbf{Challenges with coordination.}
Coordination in DM (Fig.~\ref{fig:coord}a) is hard to model in the
AM algebra because the supertag for ``and'' would need to consist only of a single
\el{and\_c} edge. We group the \el{and\_c} edge with its target node
(\textit{Mary}), creating extra supertags e.g.\ for coordinated and
non-coordinated nouns.

In PSD, coordinated arguments (\textit{John and Mary} in
Fig.~\ref{fig:coord}b) have an edge into each conjunct. This too is
hard to model with the AM algebra because after building \textit{John
  and Mary}, there can only be one node (the root source) where edges
can be attached. We therefore rewrite the graph as shown in
Fig.~\ref{fig:coord}c in preprocessing and revert the transformation
in postprocessing.

\begin{table*}[t]
	\resizebox{\linewidth}{!}{
	\begin{tabular}{lllllllllll}
	\toprule
		& \multicolumn{2}{c}{\textbf{DM}} & \multicolumn{2}{c}{\textbf{PAS}} & \multicolumn{2}{c}{ \textbf{PSD} } & \multicolumn{2}{c}{ \textbf{EDS }} & \textbf{AMR 2015} & \textbf{AMR 2017} \\
		& id F & ood F & id F & ood F & id F & ood F & Smatch F & EDM & Smatch F & Smatch F \\
		\midrule
		\citet{\wirACL} & - & - & - & - & - & - & - & - & 70.2 & 71.0 \\
		\citet{LyuTitov18AMR} & - & - & - & - & - & - & - & - & 73.7 & 74.4 \cin{0.16} \\
		\citet{Zhang19} & - & - & - & - & - & - & - & - & - & \textbf{76.3} \cin{0.1} \\
		\citet{peng17:_deep_multit_learn_seman_depen_parsin} Basic & 89.4 & 84.5 & 92.2 & 88.3 & 77.6 & 75.3 & - & - & - & - \\
		\citet{Dozat18SemanticDependency} & 93.7 & 88.9 & 94.0 & 90.8 & 81.0 & 79.4 & - & - & - & - \\
		\citet{BuysBlunsom17} & - & - & - & - & - & - & 85.5 & 85.9 & 60.1 & - \\
                \citet{chen-etal-2018-accurate}   & - & - & - & - & - & - & \textbf{90.9}\footnotemark$^,$\footnotemark & \textbf{90.4}$^1$ & - & - \\[1.0ex]             
		This paper (GloVe) & 90.4 \cin{0.2} & 84.3 \cin{0.2} & 91.4 \cin{0.1} & 86.6 \cin{0.1} & 78.1 \cin{0.2} & 74.5 \cin{0.2} & 87.6 \cin{0.1} & 82.5 \cin{0.1} & 69.2 \cin{0.4} & 70.7 \cin{ 0.2} \\
		This paper (BERT) & \textbf{93.9} \cin{0.1} & \textbf{90.3} \cin{0.1} &\textbf{94.5} \cin{0.1} & \textbf{92.5} \cin{0.1} & \textbf{82.0} \cin{0.1} &\textbf{81.5} \cin{0.3} & 90.1 \cin{0.1} & 84.9 \cin{0.1} & \textbf{74.3} \cin{0.2} & 75.3 \cin{0.2} \\
		\midrule
		\citet{peng17:_deep_multit_learn_seman_depen_parsin} Freda1 & 90.0 & 84.9 & 92.3 & 88.3 & 78.1 & 75.8 & - & - & - & - \\
        \citet{peng17:_deep_multit_learn_seman_depen_parsin} Freda3 & 90.4 & 85.3 & 92.7 & 89.0 & 78.5 & 76.4 & - & - & - & - \\ [1.0ex]
		This paper, MTL (GloVe)  & 91.2 \cin{ 0.1 } & 85.7 \cin{ 0.0 } & 92.2 \cin{ 0.2 } & 88.0 \cin{ 0.3 } & 78.9 \cin{ 0.3 } & 76.2 \cin{ 0.4 } & 88.2 \cin{ 0.1 } & 83.3 \cin{ 0.1 } &  (70.4)\footnotemark\ \cin{ 0.2} & 71.2 \cin{ 0.2} \\
		
        This paper, MTL (BERT) & \textbf{94.1} \cin{0.1} & \textbf{90.5} \cin{0.1} & \textbf{94.7} \cin{0.1} & \textbf{92.8} \cin{0.1} & \textbf{82.1} \cin{0.2} & \textbf{81.6} \cin{0.1} & 90.4 \cin{0.1} & 85.2 \cin{0.1} & (74.5)$^3$ \cin{0.1} & 75.3 \cin{0.1} \\
        \bottomrule
	\end{tabular}}
\caption{Semantic parsing accuracies (id = in domain test set; ood = out of domain test set).}
\vspace{-10pt}
\label{tab:st}
\end{table*}


\textbf{Non-decomposable graphs.} While some encodings of graphs as trees are lossy \cite{trimming-15}, ours is not: when we obtain an AM
dependency tree from a graph, that dependency tree evaluates uniquely
to the original graph.
However, not every graph in the training data can be decomposed into
an AM dependency tree in the way described above.
We mitigate the problem by making DM, PAS, and PSD graphs that have
multiple roots connected by adding an artificial root node, and by
removing \el{R-HNDL} and \el{L-HNDL} edges from EDS (2.3\% of
edges). We remove some reentrant edges in AMR as described in
Groschwitz et al.

We remove the remaining non-decomposable graphs from the training
data: 8\% of instances in DM, 6\% each for PAS and PSD, 24\% for EDS,
and 10\% for AMR. The high percentage of non-decomposable graphs in
EDS stems from the fact that EDS can align multiple nodes to the same
token, creating multi-node constants.
If more than one of these nodes are arguments or are modified
in the graph, this cannot be easily represented with the AM algebra,
and thus no valid AM dependency tree is available.

We do not
remove graphs from the test data.



\section{Evaluation} \label{sec:evaluation}

\textbf{Data.} We evaluate on the DM, PAS and PSD corpora of the
SemEval 2015 shared task \citep{OepenKMZCFHU15}, the EDS corpus
\citep{OpenSDP} and the releases LDC2015E86 and LDC2017T10 of the
AMRBank.  All corpora are named entity tagged using Stanford
CoreNLP. When tokenization, POS tags and lemmas are provided with the
data (DM, PAS, PSD), we use those. Otherwise we employ CoreNLP. We use the same
hyperparameters for all graphbanks, as detailed in the
appendix.

\textbf{Parser.} We use the BiLSTM-based arc-factored dependency
parsing model of \citet{kiperwasser16:_simpl_accur_depen_parsin_using}.
On the edge existence scores we use the hinge loss of the original K\&G model,
but we use cross-entropy loss on the edge label predictions;
this improved the accuracy of our parser. We also experimented with
the dependency parsing model of
\newcite{dozat17:deep_biaffine}, but this yielded
lower accuracies than the K\&G model.

We feed each word's BiLSTM encoding into an MLP with one hidden layer to predict
the supertags. We use separate BiLSTMs
for the dependency parser and the supertagger but share embeddings.
For every token, the BiLSTMs are fed a word embedding, the lemma, POS,
and named entity tag. In the basic version of our experiments, we used
pretrained GloVe embeddings \cite{pennington2014glove} along with trainable embeddings. In the other version we replace them by
pretrained BERT embeddings \cite{devlin2018bert}.

AMR and EDS use node labels which are nontrivially related to the
words. Therefore, we split each of their supertags into a \emph{delexicalized
  supertag} and a lexical label. For instance, instead of predicting
the supertag \G{want} in Fig.~\ref{fig:amr-am:consts} in its entirety,
we predict the label \nl{want-01} separately from the rest of the
graph. We complement the neural label prediction with a copy function
based on the word form and lemma (see supplementary materials).

 We implemented this model and Groschwitz et al.'s fixed-tree decoder within the AllenNLP framework
\cite{Gardner2017ADS}. Our code is available at \href{https://github.com/coli-saar/am-parser}{https://github.com/coli-saar/am-parser}.

\textbf{Results.} Table \ref{tab:st} (upper part) shows the results of
our basic semantic parser (with GloVe embeddings) on all six
graphbanks (mean scores over five runs and standard deviations). Our
results are competitive across the board, and set a new state of the
art for EDS Smatch scores \cite{CaiK13} among EDS parsers which are not trained on gold
syntax information. Our EDM score \cite{dridan2011parser} on EDS is lower, partially because
EDM evaluates the parser's ability to align nodes with multi-token
spans; our supertagger can only align nodes with individual tokens,
and we add alignment spans heuristically.

To test the impact of the grouping and source-naming heuristics from
Section~\ref{sec:our-decomp-meth}, we experimented with randomized
heuristics on DM. The F-score dropped by up to 18 points.

\textbf{BERT.} The use of BERT embeddings is highly effective across
the board. We set a new state of the art (without gold syntax)
on all graphbanks except AMR 2017; note that \newcite{Zhang19} also
use BERT. The improvement is particularly pronounced in the
out-of-domain evaluations, illustrating BERT's ability to transfer
across domains.

\footnotetext[1]{Uses gold syntax information from the
  HPSG DeepBank annotations at training time.}
\footnotetext[2]{Weiwei Sun, p.c.}
\footnotetext[3]{Not comparable to other AMR 2015 results because training data contained AMR 2017.}

\textbf{Multi-task learning.} Multi-task learning has been shown to
substantially improve accuracy on various semantic parsing tasks
\cite{Stanovsky18,Hershcovich18,Peng18JointDisjoint}. It is
particularly easy to apply here, because we have converted all graphbanks into a uniform format (supertags and AM dependency
trees).

We explored several multi-task approaches during development,
namely Freda
\cite{DaumeEasy07,peng17:_deep_multit_learn_seman_depen_parsin}, the
Freda generalization of
\newcite{lu16:_gener_regul_framew_domain_adapt} and the method of
\newcite{Stymne18}.
We found Freda to work best and use it
for evaluation.  Our setup compares most directly to Peng et al.'s
``Freda1'' model, concatenating the output of a graphbank-specific
BiLSTM with that of a shared BiLSTM, using graphbank-specific MLPs for
supertags and edges, and sharing input embeddings.

We pooled all corpora into a multi-task training set except for AMR 2015, since it is a subset of AMR 2017.
We also added the English Universal Dependency treebanks \citep{UD2.3}
to our training set (without any supertags).
The results on the test dataset are shown in Table~\ref{tab:st} (bottom). 
With GloVe, multi-task learning led to substantial improvements; with BERT the improvements are smaller but still noticeable.


\section{Conclusion} \label{sec:conclusion}

We have shown how to perform accurate semantic parsing across
a diverse range of graphbanks. We achieve this by training a
compositional neural parser on graphbank-specific tree decompositions
of the annotated graphs and combining it with BERT and multi-task
learning.

In the future, we would like to extend our approach to sembanks which
are annotated with different types of semantic representation, e.g.\
SQL \cite{yu18:_spider} or DRT
\cite{abzianidze-EtAl:2017:EACLshort}. Furthermore, one limitation of
our approach is that the latent AM dependency trees are determined by
heuristics, which must be redeveloped for each new
graphbank. We will explore latent-variable models to learn the
dependency trees automatically.

\section*{Acknowledgements}
We thank Stephan Oepen, Weiwei Sun and Meaghan Fowlie for helpful discussions and the reviewers for their insightful comments. This work was supported by DFG grant KO 2916/2-2.


\bibliography{mybib}

\begin{thebibliography}{32}
\expandafter\ifx\csname natexlab\endcsname\relax\def\natexlab#1{#1}\fi

\bibitem[{Abzianidze et~al.(2017)Abzianidze, Bjerva, Evang, Haagsma, van Noord,
  Ludmann, Nguyen, and Bos}]{abzianidze-EtAl:2017:EACLshort}
Lasha Abzianidze, Johannes Bjerva, Kilian Evang, Hessel Haagsma, Rik van Noord,
  Pierre Ludmann, Duc-Duy Nguyen, and Johan Bos. 2017.
\newblock \href {http://aclweb.org/anthology/E17-2039} {The parallel meaning
  bank: Towards a multilingual corpus of translations annotated with
  compositional meaning representations}.
\newblock In \emph{Proceedings of the 15th Conference of the European Chapter
  of the Association for Computational Linguistics}.

\bibitem[{Agi\'{c} et~al.(2015)Agi\'{c}, Koller, and Oepen}]{trimming-15}
\v{Z}eljko Agi\'{c}, Alexander Koller, and Stephan Oepen. 2015.
\newblock Semantic dependency graph parsing using tree approximations.
\newblock In \emph{Proceedings of the 14th International Conference on
  Computational Semantics (IWCS)}.

\bibitem[{Banarescu et~al.(2013)Banarescu, Bonial, Cai, Georgescu, Griffitt,
  Hermjakob, Knight, Koehn, Palmer, and Schneider}]{amBanarescuBCGGHKKPS13}
Laura Banarescu, Claire Bonial, Shu Cai, Madalina Georgescu, Kira Griffitt, Ulf
  Hermjakob, Kevin Knight, Philipp Koehn, Martha Palmer, and Nathan Schneider.
  2013.
\newblock \href {http://aclweb.org/anthology/W13-2322} {{A}bstract {M}eaning
  {R}epresentation for {Sembanking}}.
\newblock In \emph{Proceedings of the 7th Linguistic Annotation Workshop and
  Interoperability with Discourse}.

\bibitem[{Buys and Blunsom(2017)}]{BuysBlunsom17}
Jan Buys and Phil Blunsom. 2017.
\newblock \href {https://doi.org/10.18653/v1/P17-1112} {Robust incremental
  neural semantic graph parsing}.
\newblock In \emph{Proceedings of the 55th Annual Meeting of the Association
  for Computational Linguistics}.

\bibitem[{Cai and Knight(2013)}]{CaiK13}
Shu Cai and Kevin Knight. 2013.
\newblock Smatch: an evaluation metric for semantic feature structures.
\newblock In \emph{Proceedings of the 51st Annual Meeting of the Association
  for Computational Linguistics}.

\bibitem[{Chen et~al.(2018)Chen, Sun, and Wan}]{chen-etal-2018-accurate}
Yufei Chen, Weiwei Sun, and Xiaojun Wan. 2018.
\newblock \href {https://www.aclweb.org/anthology/P18-1038} {Accurate
  {SHRG}-based semantic parsing}.
\newblock In \emph{Proceedings of the 56th Annual Meeting of the Association
  for Computational Linguistics (Volume 1: Long Papers)}, pages 408--418,
  Melbourne, Australia. Association for Computational Linguistics.

\bibitem[{Copestake et~al.(2005)Copestake, Flickinger, Pollard, and
  Sag}]{copestake2005mrs}
Ann Copestake, Dan Flickinger, Carl Pollard, and Ivan~A Sag. 2005.
\newblock Minimal recursion semantics: An introduction.
\newblock \emph{Research on language and computation}, 3(2-3):281--332.

\bibitem[{{Daum\'e III}(2007)}]{DaumeEasy07}
Hal {Daum\'e III}. 2007.
\newblock \href {http://aclweb.org/anthology/P07-1033} {Frustratingly easy
  domain adaptation}.
\newblock In \emph{Proceedings of the 45th Annual Meeting of the Association of
  Computational Linguistics}.

\bibitem[{Devlin et~al.(2019)Devlin, Chang, Lee, and
  Toutanova}]{devlin2018bert}
Jacob Devlin, Ming-Wei Chang, Kenton Lee, and Kristina Toutanova. 2019.
\newblock \href {https://www.aclweb.org/anthology/N19-1423} {{BERT}:
  Pre-training of deep bidirectional transformers for language understanding}.
\newblock In \emph{Proceedings of the 2019 Conference of the North {A}merican
  Chapter of the Association for Computational Linguistics: Human Language
  Technologies}.

\bibitem[{Dozat and Manning(2017)}]{dozat17:deep_biaffine}
Timothy Dozat and Christopher~D. Manning. 2017.
\newblock Deep biaffine attention for neural dependency parsing.
\newblock In \emph{ICLR}.

\bibitem[{Dozat and Manning(2018)}]{Dozat18SemanticDependency}
Timothy Dozat and Christopher~D. Manning. 2018.
\newblock \href {http://aclweb.org/anthology/P18-2077} {Simpler but more
  accurate semantic dependency parsing}.
\newblock In \emph{Proceedings of the 56th Annual Meeting of the Association
  for Computational Linguistics}.

\bibitem[{Dridan and Oepen(2011)}]{dridan2011parser}
Rebecca Dridan and Stephan Oepen. 2011.
\newblock Parser evaluation using elementary dependency matching.
\newblock In \emph{Proceedings of the 12th International Conference on Parsing
  Technologies}, pages 225--230.

\bibitem[{Flickinger et~al.(2017)Flickinger, Haji{\v c}, Ivanova, Kuhlmann,
  Miyao, Oepen, and Zeman}]{OpenSDP}
Dan Flickinger, Jan Haji{\v c}, Angelina Ivanova, Marco Kuhlmann, Yusuke Miyao,
  Stephan Oepen, and Daniel Zeman. 2017.
\newblock \href {http://hdl.handle.net/11234/1-1956} {Open {SDP} 1.2}.
\newblock {LINDAT}/{CLARIN} digital library at the Institute of Formal and
  Applied Linguistics ({{\'U}FAL}), Faculty of Mathematics and Physics, Charles
  University.

\bibitem[{Gardner et~al.(2017)Gardner, Grus, Neumann, Tafjord, Dasigi, Liu,
  Peters, Schmitz, and Zettlemoyer}]{Gardner2017ADS}
Matthew Gardner, Joel Grus, Mark Neumann, Oyvind Tafjord, Pradeep Dasigi,
  Nelson H~S Liu, Matthew Peters, Michael Schmitz, and Luke~S. Zettlemoyer.
  2017.
\newblock A deep semantic natural language processing platform.

\bibitem[{Groschwitz(2019)}]{GroschwitzDiss}
Jonas Groschwitz. 2019.
\newblock \href {http://www.coli.uni-saarland.de/~jonasg/thesis.pdf}
  {\emph{Methods for taking semantic graphs apart and putting them back
  together again}}.
\newblock Ph.D. thesis, Macquarie University and Saarland University.

\bibitem[{Groschwitz et~al.(2017)Groschwitz, Fowlie, Johnson, and
  Koller}]{graph-algebra-17}
Jonas Groschwitz, Meaghan Fowlie, Mark Johnson, and Alexander Koller. 2017.
\newblock \href {http://aclweb.org/anthology/W17-6810} {A constrained graph
  algebra for semantic parsing with {AMRs}}.
\newblock In \emph{Proceedings of the 12th International Conference on
  Computational Semantics (IWCS)}.

\bibitem[{Groschwitz et~al.(2018)Groschwitz, Lindemann, Fowlie, Johnson, and
  Koller}]{groschwitz18:_amr_depen_parsin_typed_seman_algeb}
Jonas Groschwitz, Matthias Lindemann, Meaghan Fowlie, Mark Johnson, and
  Alexander Koller. 2018.
\newblock \href {http://aclweb.org/anthology/P18-1170} {A{MR} {Dependency
  Parsing with a Typed Semantic Algebra}}.
\newblock In \emph{Proceedings of ACL}.

\bibitem[{Hershcovich et~al.(2018)Hershcovich, Abend, and
  Rappoport}]{Hershcovich18}
Daniel Hershcovich, Omri Abend, and Ari Rappoport. 2018.
\newblock \href {http://aclweb.org/anthology/P18-1035} {Multitask parsing
  across semantic representations}.
\newblock In \emph{Proceedings of the 56th Annual Meeting of the Association
  for Computational Linguistics}.

\bibitem[{Kiperwasser and
  Goldberg(2016)}]{kiperwasser16:_simpl_accur_depen_parsin_using}
Eliyahu Kiperwasser and Yoav Goldberg. 2016.
\newblock \href {http://aclweb.org/anthology/Q16-1023} {S{imple and Accurate
  Dependency Parsing Using Bidirectional LSTM Feature Representations}}.
\newblock \emph{Transactions of the Association for Computational Linguistics},
  4:313--327.

\bibitem[{Kuhlmann and
  Oepen(2016)}]{kuhlmann16:_towar_catal_linguis_graph_banks}
Marco Kuhlmann and Stephan Oepen. 2016.
\newblock \href {http://aclweb.org/anthology/J16-4009} {Towards a catalogue of
  linguistic graph banks}.
\newblock \emph{Computational Linguistics}, 42(4):819--827.

\bibitem[{Lu et~al.(2016)Lu, Chieu, and
  L{\"o}fgren}]{lu16:_gener_regul_framew_domain_adapt}
Wei Lu, Hai~Leong Chieu, and Jonathan L{\"o}fgren. 2016.
\newblock A general regularization framework for domain adaptation.
\newblock In \emph{Proceedings of EMNLP}.

\bibitem[{Lyu and Titov(2018)}]{LyuTitov18AMR}
Chunchuan Lyu and Ivan Titov. 2018.
\newblock \href {http://aclweb.org/anthology/P18-1037} {A{MR Parsing as Graph
  Prediction with Latent Alignment}}.
\newblock In \emph{Proceedings of the 56th Annual Meeting of the Association
  for Computational Linguistics}.

\bibitem[{Manning et~al.(2014)Manning, Surdeanu, Bauer, Finkel, Bethard, and
  McClosky}]{ManningSBFBM14}
Christopher~D. Manning, Mihai Surdeanu, John Bauer, Jenny Finkel, Steven~J.
  Bethard, and David McClosky. 2014.
\newblock The stanford corenlp natural language processing toolkit.
\newblock In \emph{Proceedings of 52nd Annual Meeting of the Association for
  Computational Linguistics: System Demonstrations}.

\bibitem[{Nivre et~al.(2018)Nivre, Abrams, Agi{\'c}, Ahrenberg, Antonsen,
  Aplonova, Aranzabe et~al.}]{UD2.3}
Joakim Nivre, Mitchell Abrams, {\v Z}eljko Agi{\'c}, Lars Ahrenberg, Lene
  Antonsen, Katya Aplonova, Maria~Jesus Aranzabe, et~al. 2018.
\newblock \href {http://hdl.handle.net/11234/1-2895} {Universal dependencies
  2.3}.
\newblock {LINDAT}/{CLARIN} digital library at the Institute of Formal and
  Applied Linguistics ({{\'U}FAL}), Faculty of Mathematics and Physics, Charles
  University.

\bibitem[{Oepen et~al.(2015)Oepen, Kuhlmann, Miyao, Zeman, Cinkov\'{a},
  Flickinger, Haji\v{c}, and Ure\v{s}ov\'{a}}]{OepenKMZCFHU15}
Stephan Oepen, Marco Kuhlmann, Yusuke Miyao, Daniel Zeman, Silvie Cinkov\'{a},
  Dan Flickinger, Jan Haji\v{c}, and Zde\v{n}ka Ure\v{s}ov\'{a}. 2015.
\newblock \href {http://aclweb.org/anthology/S15-2153} {Semeval 2015 task 18:
  Broad-coverage semantic dependency parsing}.
\newblock In \emph{Proceedings of the 9th International Workshop on Semantic
  Evaluation (SemEval 2015)}.

\bibitem[{Peng et~al.(2017)Peng, Thomson, and
  Smith}]{peng17:_deep_multit_learn_seman_depen_parsin}
Hao Peng, Sam Thomson, and Noah~A. Smith. 2017.
\newblock \href {http://aclweb.org/anthology/P17-1186} {Deep {Multitask
  Learning for Semantic Dependency Parsing}}.
\newblock In \emph{Proceedings of ACL}.

\bibitem[{Peng et~al.(2018)Peng, Thomson, Swayamdipta, and
  Smith}]{Peng18JointDisjoint}
Hao Peng, Sam Thomson, Swabha Swayamdipta, and Noah~A. Smith. 2018.
\newblock \href {https://doi.org/10.18653/v1/N18-1135} {Learning joint semantic
  parsers from disjoint data}.
\newblock In \emph{Proceedings of the 2018 Conference of the North American
  Chapter of the Association for Computational Linguistics: Human Language
  Technologies}, pages 1492--1502.

\bibitem[{Pennington et~al.(2014)Pennington, Socher, and
  Manning}]{pennington2014glove}
Jeffrey Pennington, Richard Socher, and Christopher~D. Manning. 2014.
\newblock Glove: Global vectors for word representation.
\newblock In \emph{Empirical Methods in Natural Language Processing (EMNLP)}.

\bibitem[{Stanovsky and Dagan(2018)}]{Stanovsky18}
Gabriel Stanovsky and Ido Dagan. 2018.
\newblock \href {http://aclweb.org/anthology/D18-1263} {Semantics as a foreign
  language}.
\newblock In \emph{Proceedings of the 2018 Conference on Empirical Methods in
  Natural Language Processing}.

\bibitem[{Stymne et~al.(2018)Stymne, de~Lhoneux, Smith, and Nivre}]{Stymne18}
Sara Stymne, Miryam de~Lhoneux, Aaron Smith, and Joakim Nivre. 2018.
\newblock \href {http://aclweb.org/anthology/P18-2098} {Parser training with
  heterogeneous treebanks}.
\newblock In \emph{Proceedings of the 56th Annual Meeting of the Association
  for Computational Linguistics}.

\bibitem[{Yu et~al.(2018)Yu, Zhang, Yang, Yasunaga, Wang, Li, Ma, Li, Yao,
  Roman, Zhang, and Radev}]{yu18:_spider}
Tao Yu, Rui Zhang, Kai Yang, Michihiro Yasunaga, Dongxu Wang, Zifan Li, James
  Ma, Irene Li, Qingning Yao, Shanelle Roman, Zilin Zhang, and Dragomir Radev.
  2018.
\newblock \href {http://aclweb.org/anthology/D18-1425} {Spider: A large-scale
  human-labeled dataset for complex and cross-domain semantic parsing and
  text-to-sql task}.
\newblock In \emph{Proceedings of the 2018 Conference on Empirical Methods in
  Natural Language Processing}.

\bibitem[{Zhang et~al.(2019)Zhang, Ma, Duh, and Durme}]{Zhang19}
Sheng Zhang, Xutai Ma, Kevin Duh, and Benjamin~Van Durme. 2019.
\newblock {AMR} parsing as sequence-to-graph transduction.
\newblock In \emph{Proceedings of the 57th Annual Meeting of the Association
  for Computational Linguistics (ACL)}.

\end{thebibliography}
\bibliographystyle{acl_natbib}

\appendix

\section{Edge Attachment and Source Heuristics}

This section presents details of the heuristics discussed in Section~3 of the main paper, concerning grouping (edge attachment) and source heuristics.

Tables \ref{tab:dmh}, \ref{tab:pash}, \ref{tab:psdh} and \ref{tab:edsh} show our edge attachment and source assignment heuristics for DM, PAS, PSD and EDS respectively. The heuristics are broken down by the edge's label in the `Label' column (`*' is a wildcard matching any string). A checkmark (\checkmark) in the `To Origin' column means that all edges with this label are attached to their origin node, a cross (\xmark) means the edge is attached to its target node. In EDS, all edges go with their origin in principle, but in order to improve decomposability, we attach an edge to its target if the target node has label \textit{udef\_q} or \textit{nominalization}.

The graphbanks differ in the directionality of edges; in particular, modifier relations sometimes point from the head to the modifier (PSD, AMR) and sometimes from the modifier to the head (DM, PAS, EDS). Our edge assignment heuristics account for that, following the principles for grouping. In DM, for instance, we treat the BV edge (pointing from a determiner to its head noun) as a modifier edge, and thus, it belongs to the determiner, which happens to be at the origin of the edge.

The `Source' column specifies which source is assigned to an empty node attached to an as-graph, depending on the label of the edge with which the node is attached. If an as-graph has multiple attached edges with the same label (or labels that map to the same source), i.e.~multiple nodes would obtain the same source, we disambiguate the sources by sorting the nodes with the same source in an arbitrary order and appending `2' to the source at the second node, `3' to the source at the third node and so on (the source at the first node remains unchanged).
In PSD, where this happens particularly often, we order the nodes with the same source in their word order rather than arbitrarily, to get more consistent AM dependency trees.
For example, if in PSD there are two nodes that are attached to an as-graph with \el{CONJ.member} edges (such as in Figure~\ref{fig:coord}c in the main paper), the edge going to the left gets assigned an \src{op} source and the edge going to the right an \src{op2} source.

\paragraph{Passive and object promotion.} Following
\citet{groschwitz18:_amr_depen_parsin_typed_seman_algeb}, we allow some source names to be changed or swapped in an as-graph constant after their original assignments. That is, after we build a constant according to the edge grouping and source assignments described above, we generate multiple variants of the constant that have different source names. We allow
\begin{itemize}
    \item object promotion, e.g.~instead of an \src{O3} source we may
      also use an \src{O2} or \src{O} source, as long as they don't
      exist yet in the constant,
    \item unaccusative subjects, i.e.~an \src{O} source may become an \src{S} source if no \src{S} source is present yet in the constant, and
    \item passive, i.e.~switching \src{O} and \src{S} sources.
\end{itemize}
This allows more graphs to be decomposed, by allowing e.g.~the coordination of a verb in active and a verb in passive, or the raising of unaccusative subjects. We also follow \citet{groschwitz18:_amr_depen_parsin_typed_seman_algeb} in the following (quoted directly from their Section 4.2): ``To make our as-graphs more consistent, we prefer constants that promote objects as far as possible, use unaccusative subjects, and no passive alternation, but still allow constants that do not satisfy these conditions if necessary.''

\begin{table}
	\small
	\centering
	\begin{tabular}{lll}
		\toprule
		\textbf{Label} & \textbf{To Origin} & \textbf{Source} \\
		ARG1 & \checkmark & S \\
		ARG2 & \checkmark   & O  \\
		compound & \checkmark   & comp \\
		BV & \checkmark   & D \\
		poss & \checkmark  & poss  \\
		\_*\_c & \xmark  & coord\\
		ARG3 & \checkmark  & O2 \\
		mwe & \xmark & comp \\
		conj & \xmark & coord \\
		plus & \xmark & M  \\
		ARG4 & \checkmark  & O3\\
		all other edges & \checkmark  & M\\
		\bottomrule
	\end{tabular}
	\caption{Heuristics for DM. }
	\label{tab:dmh}
\end{table}
\begin{table}
	\small
	\centering
	\begin{tabular}{lll}
		\toprule
		\textbf{Edge label} & \textbf{To Origin} & \textbf{Source} \\
		det\_ARG1 & $\checkmark$& D \\
		punct\_ARG1 & $\checkmark$& pnct \\
		coord\_ARG$_i$ & $\checkmark$& op$_i$ \\
		verb\_ARG1 & $\checkmark$& S \\
		*\_ARG1 & $\checkmark$& O \\
		*\_ARG2 & $\checkmark$& O \\
		all other edges& $\checkmark$ & M \\
		\bottomrule
	\end{tabular}
	\caption{Heuristics for PAS.}
	\label{tab:pash}
\end{table}

\paragraph{Reentrancy heuristics.} We update the reentrancy patterns of \citet{graph-algebra-17} in the following way. No coordination node patterns are allowed in DM (since DM uses edges for coordination); Coordination nodes in PAS are characterized via their coord\_ARG$_i$ edges. In PSD and EDS, any node that has two arguments that themselves have a common argument can be a coordination node.

Raising in PAS is done with the coordination pattern; in the others, a node $v$ where one argument $w$ has an \src{S} source can be a raising node (that is, we add an $[\src{S}]$-annotation at the source that the $v$-constant has at node $w$), as long as the edge between $v$ and $w$ has label
\begin{itemize}
    \item ARG1 or ARG2 in DM,
    \item PAT-arg in PSD, or
    \item any label in EDS.
\end{itemize}
We use the same \sortof{raising}-style pattern for comparatives in PSD, where we use no condition on the source that is \sortof{passed along}, but the edge from $v$ to $w$ must have label \el{CPR}.

\begin{table}
	\small
	\centering
	\begin{tabular}{lll}
		\toprule
		\textbf{Label} & \textbf{To Origin} & \textbf{Source} \\
		ACT-arg & $\checkmark$ & S \\
		PAT-arg & $\checkmark$ & O \\
		*-arg (except ACT, PAT) & $\checkmark$ & OO \\
		*.member & $\checkmark$ & op \\
		CPR & $\checkmark$ & M \\
		all other edges & \xmark & M \\
		\bottomrule
	\end{tabular}
	\caption{Heuristics for PSD.}
	\label{tab:psdh}
\end{table}

\begin{table}
	\small
	\centering
	\begin{tabular}{lll}
		\toprule
		\textbf{Label}  &  \textbf{To Origin} & \textbf{Source}\\
		ARG1 & $\checkmark$ & S \\
		ARG2 & $\checkmark$ & O  \\
		BV & $\checkmark$ & D \\
		R-INDEX & $\checkmark$ &  op1 \\
		L-INDEX & $\checkmark$ & op2 \\
		R-HNDL & $\checkmark$ & op1  \\
		L-HNDL & $\checkmark$ &  op2  \\
		ARG* & $\checkmark$ &  O \\
		all other edges & $\checkmark$ & M \\
		\bottomrule
	\end{tabular}
	\caption{Heuristics for EDS.}
	\label{tab:edsh}
\end{table}

\paragraph{Randomized heuristics.} The randomized heuristics we experimented with on the DM set choose edge grouping (to target or to origin) and source names for each edge label independently uniformly at random (but consistently across the corpus).
\section{Training and Parsing Details}
\label{sec:hyperparams}

We reimplemented the graph-based parser of \citet{kiperwasser16:_simpl_accur_depen_parsin_using} in AllenNLP. We deviate from the original implementation in the following:
\begin{itemize}
	\item We use a cross-entropy loss instead of a hinge loss on the edge \textit{label} predictions.
	\item We follow \citet{\wirACL} in using the Chu-Liu-Edmonds algorithm instead of Eisner's algorithm.
	\item We don't perform word dropout but regular dropout on the input.
\end{itemize}
We add a supertagger consisting of a separate BiLSTM, from whose states we predict delexicalized graph fragments and lexical labels with an MLP. Learned embeddings are shared between the BiLSTM of the supertagger and the dependency parser.

The hyperparameters are collected in table \ref{tab:hyper}. We train the parser for 40 epochs and pick the model with the highest performance on the development set (measured in Smatch for EDS, not in EDM). We perform early stopping with patience of 10 epochs.
Every lemma (and word in the case of using GloVe embeddings) that occurs fewer than 7 times is treated as unknown.

We use BucketIterators (padding noise 0.1) and the methods implemented in AllenNLP for performing padding and masking.

\paragraph{Training with GloVe} We use the 200-dimensional version of GloVe (6B.200d) along with 100-dimensional trainable embeddings. We use two layers in the BiLSTMs and train with a batch size of 48.

\paragraph{Training with BERT}  When using BERT, we replace both the GloVe embeddings and the learned word embedding with BERT. Since BERT does not
provide embeddings for the artificial root of the dependency tree, we learn a separate embedding. In some graphbanks (DM, PAS, PSD), we also have an
artificial word at the end of each sentence, that is used to connect the graphs. From BERT's perspective, the artificial word is a period symbol.

When training with BERT, we use a batch size of 64 and only one layer in the BiLSTMs. We use the "large-uncased" model as available through AllenNLP and don't fine-tune BERT.

\begin{table}
	\small
	\begin{tabular}{ll}
		\toprule
		Activation function & tanh \\
		Optimizer & Adam \\
		Learning rate & 0.001 \\
		Epochs & 40 \\
		\midrule
		Dim of lemma embeddings & 64 \\
		Dim of POS embeddings & 32 \\
		Dim of NE embeddings & 16 \\
		Minimum lemma frequency & 7 \\
		\midrule
		Hidden layers in all MLPs & 1 \\
		\midrule
		Hidden units in LSTM (per direction) & 256 \\
		Hidden units in edge existence MLP & 256 \\
		Hidden units in edge label MLP & 256 \\
		Hidden units in supertagger MLP & 1024 \\
		Hidden units in lexical label tagger MLP & 1024 \\
		\midrule
		Layer dropout in LSTMs & 0.3 \\
		Recurrent dropout in LSTMs & 0.4 \\
		Input dropout & 0.3\\
		Dropout in edge existence MLP & 0.0 \\
		Dropout in edge label MLP & 0.0 \\
		Dropout in supertagger MLP & 0.4 \\
		Dropout in lexical label tagger MLP & 0.4 \\
		\bottomrule
	\end{tabular}
	\caption{Common hyperparameters used in all experiments.}
	\label{tab:hyper}
\end{table}

\paragraph{MTL} In our Freda experiments, we have one LSTM per graphbank and one that is shared between the graphbanks. 
When we compute scores for a sentence, we run it through its graphbank-specific LSTM and the shared one. We concatenate the outputs and feed it to graphbank-specific MLPs.
Again, we have separate LSTM for the edge model (input to edge existence and edge label MLP) and the supertagging model. 
In effect, we have two LSTMs that are shared over the graphbanks: one for the edge model and one for the supertagging model.

All LSTMs have the hyperparameters detailed in table \ref{tab:hyper}. In the case of UD, we don't use a graphbank-specific supertagger because there are no supertags for UD. We don't pool the UD treebanks together. 

In the MTL setup, we select the epoch with the highest development F-score for DM for evaluation on all test sets.
\paragraph{Parsing} We follow \citet{\wirACL} in predicting the best
unlabeled dependency tree with the Chu-Liu-Edmonds algorithm and then
run their fixed-tree decoder restricted to the 6 best supertags. This
computes the best well-typed AM dependency tree with the same shape as
the unlabeled tree. 

Parsing is usually relatively fast (between 30 seconds and 2 minutes for the test
corpora) but very slow for a few sentences very long sentences in the AMR test
corpora. Therefore, we set a timeout. If parsing with $k$ supertags is
not completed within 30 minutes, we retry with $k-1$ supertags. If
$k=0$, we use a dummy graph with a single node. This happened 4 times over different runs on AMR with the basic version of the parser and once when using BERT.

\paragraph{Copy function} In order to predict the lexical label for EDS and AMR, we predict only the difference to its lemma or word form. For instance, if the lexical label is "want-01", we try to predict \texttt{\$LEMMA\$-01} instead at the word in question, e.g. wanted, and restore the full form of the lexical label in postprocessing.

\section{Details of Preprocessing and Postprocessing}

\paragraph{DM, PAS and PSD} We handle disconnected graphs with components that contain more than one node by adding an artificial word to the end of the sentence. We draw an edge from this word to one node in every weakly connected component of the graph. We select this node by invoking Stanford CoreNLP \citep{ManningSBFBM14} to find the head of the span the component comprises. 

Disconnected components that only contain one word are treated as words without semantic contribution, which we attach to the artificial root (position 0) with an \textsc{Ignore}-edge.

Since the node labels in these graphbanks are the words of the sentences, we simply copy the words over to the graph.

We use the evaluation toolkit that was developed for the shared task: \href{https://github.com/semantic-dependency-parsing/toolkit}{https://github.com/semantic-dependency-parsing/toolkit}.

\paragraph{EDS} We only consider connected EDS graphs (98.5\% of the corpus) and follow \citet{BuysBlunsom17} regarding options for the tokenizer except for hyphenated words, which we split. Since EDS nodes are aligned with (character) spans in the sentence, we make use of this information in the decomposition. In our approach, however, we require every graph constant to stem from exactly one token. In order to enforce this, we assign nodes belonging to a multi-token span to an atomic span whose nodes are incident. For consistency, we perform this from left to right. We try to avoid creating graph constants that would require more than one root source. Where this fails, the graph cannot be decomposed.

We delete R-HNDL and L-HNDL edges only if this does not make the graph disconnected. Thus, we need heuristics for them (see table \ref{tab:edsh}).

Before delexicalizing graphs constants, we need to identify lexical nodes. A node is considered lexical if has an incoming c-arg edge or if its label is similar to the aligned word, its lemma or its modified lemma. We compute the modified lemma by a few hand-written rules from the CoreNLP lemma. For instance, "Tuesday" is mapped to "Tue". We also re-inflect adverbs (as identified by the POS tagger) to their respective adjectives if possible, e.g. "interestingly" becomes "interesting". We perform this step in order to be able to represent the lexical label of more graph constants as function of the word which they belong to. The modified lemma is not used as input to the neural network.

When performing the delexicalization, we replace the character span information with placeholders indicating if this span is atomic (comprises a single word) or not. We restore the span information for every node with a very simple heuristic in postprocessing: If the span is atomic, we simply look up the character span in the original string. For nodes with complex spans, we compute the minimum of beginnings and the maximum of endings of its children. In terms of evaluation, the span information is relevant only for EDM. Comparing the graphs that we restore from our training data to the gold standard, we find that the upper bound is at 89.7 EDM F-score. The upper bound in terms of Smatch is at 96.9 F-score.

We use EDM in an implementation by \citet{BuysBlunsom17}.

\paragraph{UD} Since UD POS tags are different from the English PTB tagset, we use CoreNLP to tag the UD treebanks. We use the English treebanks EWT, GUM, ParTUT and LinES \citep{UD2.3}.

\paragraph{AMR}
We use the pre- and postprocessing pipeline of \citet{\wirACL}. We conflate named entities in preprocessing. For instance, "New York" is conflated to one token "New\_York". When such a graph constant is predicted, we restore the named entity prior to evaluation.

\end{document}